\pdfoutput=1

\documentclass[10pt,journal]{IEEEtran}

\usepackage{amssymb}
\usepackage{amsmath}
\usepackage{graphicx}
\usepackage{tabularx}
\usepackage{times}
\usepackage{latexsym}
\usepackage{array}
\usepackage{longtable}
\usepackage{booktabs}
\usepackage{arydshln}
\usepackage{hyperref}
\usepackage{breakurl}
\usepackage{wrapfig}
\usepackage[T1]{fontenc}
\usepackage[utf8]{inputenc}
\usepackage{microtype}
\usepackage{inconsolata}
\usepackage{graphicx}
\usepackage{tabularx}
\usepackage{amsmath,amssymb,amsfonts}
\usepackage{etoolbox}
\usepackage{pifont}
\usepackage{textcomp}
\usepackage{color}
\usepackage{amssymb}
\usepackage{amsmath}
\usepackage{algorithm}
\usepackage{algorithmic}
\usepackage{subfigure}
\usepackage{multirow}
\usepackage{booktabs}
\usepackage{xcolor}
\usepackage{colortbl,booktabs}

\begin{document}

\title{Iterative Data Generation with Large Language Models for Aspect-based Sentiment Analysis}

\author{Qihuang~Zhong,~\IEEEmembership{Member,~IEEE,}
        Haiyun~Li,
        Luyao~Zhuang,
        Juhua~Liu,~\IEEEmembership{Member,~IEEE,}
        Bo~Du,~\IEEEmembership{Senior~Member,~IEEE}
\thanks{This work was supported in part by the National Key Research and Development Program of China under Grant 2023YFC2705700, in part by the National Natural Science Foundation of China under Grants 623B2076, U23B2048, 62076186 and 62225113, and in part by the Innovative Research Group Project of Hubei Province under Grant 2024AFA017. The numerical calculations in this paper have been done on the supercomputing system in the Supercomputing Center of Wuhan University. \textit{Equal contribution: Qihuang Zhong, Haiyun Li, and Luyao Zhuang contribute equally to this work. Corresponding Author: Juhua Liu, Bo Du (e-mail: \{liujuhua, dubo\}@whu.edu.cn).}}

\thanks{Q. Zhong, H. Li, L. Zhuang, J. Liu and B. Du are with the School of Computer Science, National Engineering Research Center for Multimedia Software, Institute of Artificial Intelligence, and Hubei Key Laboratory of Multimedia and Network Communication Engineering, Wuhan University, Wuhan, China (e-mail: \{zhongqihuang, haiyunli.whu, liujuhua, dubo\}@whu.edu.cn; a523550914@gmail.com).}
}

\maketitle

\begin{abstract}
Aspect-based Sentiment Analysis (ABSA) is an important sentiment analysis task, which aims to determine the sentiment polarity towards an aspect in a sentence. Due to the expensive and limited labeled data, data generation (DG) has become the standard for improving the performance of ABSA. However, current DG methods usually have some shortcomings: 1) poor fluency and coherence, 2) lack of diversity of generated data, and 3) reliance on some existing labeled data, hindering its applications in real-world scenarios. With the advancement of large language models (LLMs), LLM-based DG has the potential to solve the above issues. Unfortunately, directly prompting LLMs struggles to generate the desired pseudo-label ABSA data, as LLMs are prone to hallucinations, leading to undesired data generation. To this end, we propose a systematic \textbf{I}terative \textbf{D}ata \textbf{G}eneration framework, namely \textbf{IDG}, to boost the performance of ABSA. The core of IDG is to make full use of the powerful abilities (\textit{i.e.}, instruction-following, in-context learning and self-reflection) of LLMs to iteratively generate more fluent and diverse pseudo-label data, starting from an unsupervised sentence corpus. Specifically, IDG designs a novel iterative data generation mechanism and a self-reflection data filtering module to tackle the challenges of unexpected data generation caused by hallucinations. Extensive experiments on four widely-used ABSA benchmarks show that IDG brings consistent and significant performance gains among five baseline ABSA models. More encouragingly, the synthetic data generated by IDG can achieve comparable or even better performance against the manually annotated data.
\end{abstract}


%
\IEEEpeerreviewmaketitle

\section{Introduction}
\label{sec:intro}
\IEEEPARstart{A}{S} an important fine-grained sentiment analysis task, aspect-based Sentiment Analysis (ABSA) aims to determine the sentiment polarity towards an aspect in a sentence~\cite{liu2012survey,schouten2015survey}. With the advancements of pretrained language models (PLMs), \textit{e.g.}, BERT~\cite{devlin2018bert} and its variants~\cite{liu2019roberta,he2020deberta}, numerous PLM-based ABSA models have been proposed and achieved promising results~\cite{wang-etal-2020-relational,Zhong2022KnowledgeGA}. However, these methods usually require large-scale labeled fine-grained data, which is time-consuming and expensive for many emerging scenarios~\cite{yu2023cross}.

To alleviate this issue, a common approach is data generation (DG) that aims to enrich the training data and can be generally divided into two categories: word-level~\cite{wei-zou-2019-eda,wu2019conditional} and sentence-level DG~\cite{sennrich-etal-2016-improving,wang-etal-2022-contrastive}. Specifically, word-level DG methods augment the existing sentences by replacing or inserting words into sentences, leveraging techniques such as word synonym dictionaries~\cite{wei-zou-2019-eda} or contextual word embeddings~\cite{wu2019conditional}. Conversely, sentence-level DG methods focus on generating new sentences using paraphrasing methods~\cite{guo2019augmenting}, generative models~\cite{wang-etal-2022-contrastive}, or machine translation~\cite{sennrich-etal-2016-improving} techniques.
In general, these methods aim to introduce linguistic variations, while keeping the aspect and its sentiment polarity unchanged.

\begin{figure}[t]
    \centering   
    \includegraphics[width=0.48\textwidth]{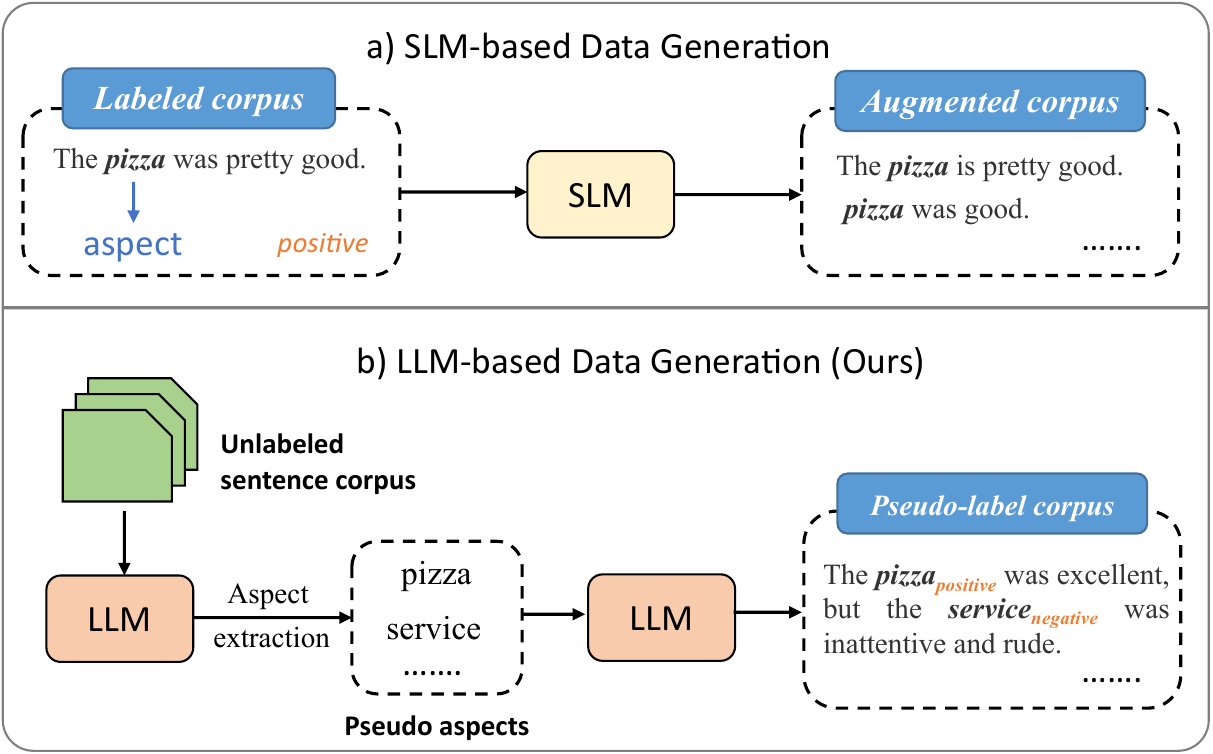}
    \caption{Comparison between our LLM-based data generation and prior small language model (SLM)-based methods. As seen, our method does not rely on the existing labeled data and can generate more high-quality and diverse pseudo-label ABSA data.}
    \label{fig:intro}
\end{figure}

Despite achieving remarkable performance, we find that the aforementioned DG methods still have some limitations: 1) \textit{Poor fluency and coherence}, as the word-level DG methods might distort the sentence meaning or structures, and current sentence-level DG methods usually struggle to generate fluent and coherent sentences~\cite{yu2023cross}. 2) \textit{Lack of the diversity of generated data}, as most of the prior DG methods do not reconstruct the structure of original sentence, limiting the diversity of generated sentences. 3) \textit{Reliance on some existing labeled data}, as these DG methods generally start from a set of existing labeled data, which could be unavailable in real-world scenarios, especially in some emerging domains. 
Intuitively, the currently popular large language models (LLMs)~\cite{openai2023gpt4,touvron2023llamav2} have the great potential to deal with the above issues of DG methods, as they have the potential to generate fluent and high-quality text following the human instructions~\cite{wei2021finetuned}. Hence, there raises a question: \textit{``whether we can leverage the powerful ability of LLMs for better augmenting the ABSA data"}?
As illustrated in Fig.~\ref{fig:intro}, compared to prior DG methods, LLM-based data generation does not rely on the supervised corpus and has the potential to generate more diverse data. Actually, there are some existing LLM-based data generation methods in the NLP field~\cite{ding2024data,bayer2023data}, but their applications on the ABSA are still under-explored. A major challenge is that LLMs are prone to hallucinations during fine-grained data generation, leading to the unstable diversity and quality of the synthesis data. 

Motivated by this, we propose a novel \textbf{I}terative \textbf{D}ata \textbf{G}eneration approach, namely \textbf{IDG}, which aims to generate fluent and diverse ABSA training data. The core of IDG is to make full use of the powerful abilities (\textit{i.e.}, instruction-following, in-context learning and self-reflection) of LLMs to improve the generated data, ensuring both diversity and quality. 
Specifically, given an easy-to-obtain unlabeled sentence corpus, IDG \ding{182} first prompts the LLM to extract the aspect terms and expand them into a candidate aspect set. Then, IDG \ding{183} introduces an iterative generation module for guiding the LLM to iteratively obtain the fluent ABSA data based on the aspect set. Lastly, to ensure the quality and diversity of the generated data, IDG \ding{184} designs a discriminator that encourages the LLM to self-reflect the synthesis data and uses the high-quality data as feedback to further guide the generation in stage~\ding{183}.
In general, the generation processes of IDG are systemic and do not rely on much existing ABSA data or human effort. That is, our IDG can be easily applied in real-world scenarios.

We evaluate our IDG on a variety of widely-used ABSA benchmarks, including Laptop14, Restaurant14~\cite{pontiki-etal-2014-semeval}, Restaurant15~\cite{pontiki-etal-2015-semeval} and Restaurant16~\cite{pontiki-etal-2016-semeval}, and the results show that: 1) our IDG brings consistent and significant performance gains among five baseline ABSA models; 2) without relying on any labeled data, IDG can achieve comparable performance to that training with full labeled data; 3) IDG outperforms the other DG counterparts by a clear margin. More in-depth analyses delve into the mechanism of IDG, and reveal when and where to use it. To summarize, our contributions are three-fold: (1) We propose a novel iterative DG approach (IDG) for ABSA by leveraging the powerful abilities (\textit{i.e.}, instruction-following, in-context learning and self-reflection) of LLMs. (2) IDG is plug-and-play and can be easily applied in real-world scenarios. (3) Extensive results on four widely-used ABSA benchmarks show the effectiveness and superiority of IDG.

The rest of this paper is organized as follows. In Sec.~\ref{sec:related}, we briefly review the related works. In Sec.~\ref{sec:method}, we introduce our proposed framework in detail. Sec.~\ref{sec:experiments} reports and discusses our experimental results. Lastly, we conclude our study in Sec.~\ref{sec:conclusion}.

\begin{figure*}[t]
    \centering   
    \includegraphics[width=0.84\textwidth]{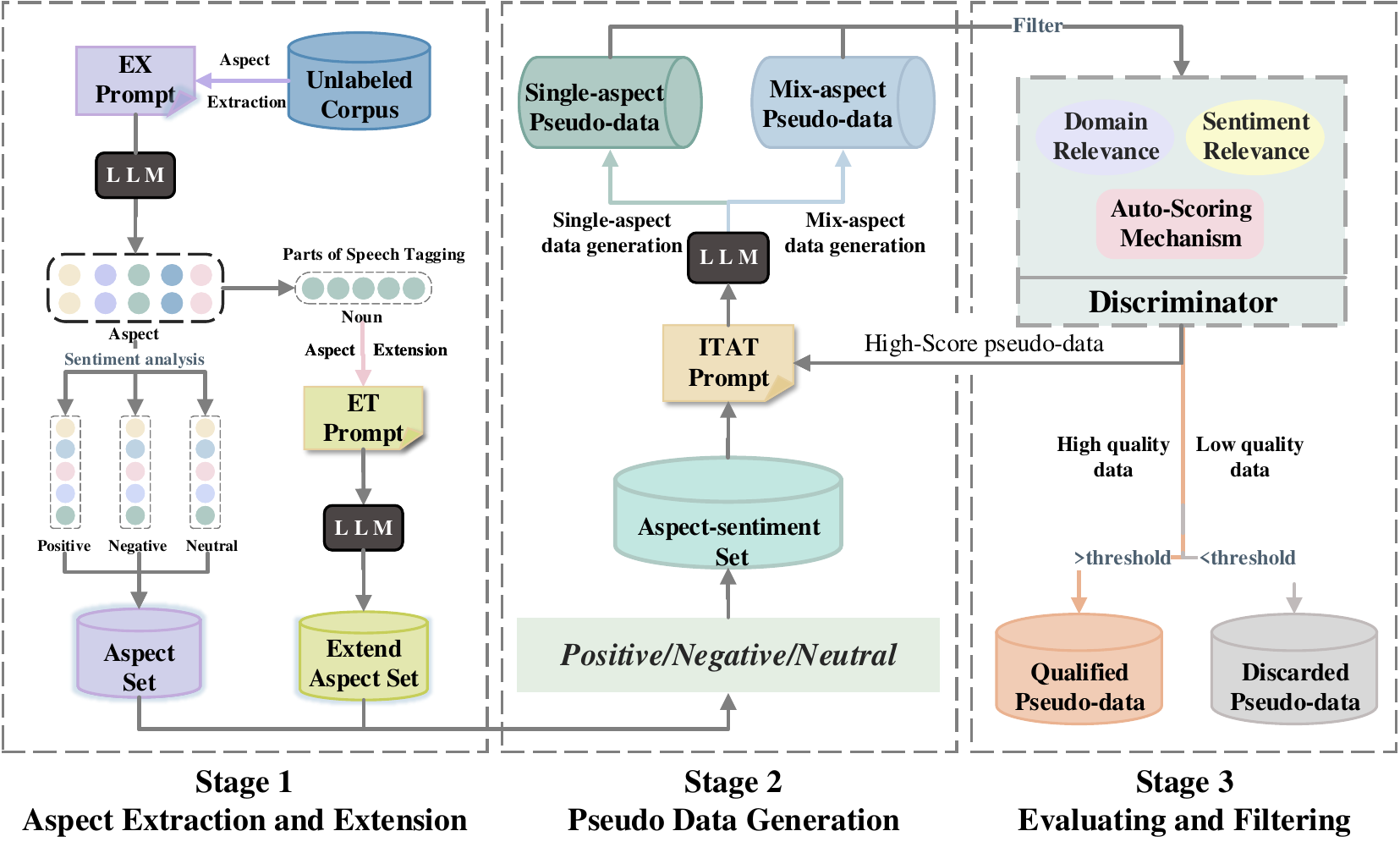}
    \caption{\textbf{Overview of our IDG framework}, covering three-stage processes: \ding{182} \textit{Aspect Extraction and Extension}, \ding{183} \textit{Pseudo Data Generation} and \ding{184} \textit{Evaluating and Filtering}. Notably, ``EX Prompt'' and ``ET Prompt'' denote the aspect extraction and extension prompts, respectively. ``ITAT Prompt'' refers to the Iteration Teaching Analysis Prompt, which enforces the LLM to generate more diverse data. 
    }
    \label{fig:IDG}
    \end{figure*}

\section{Related Work}
\label{sec:related}
\subsection{Aspect-based Sentiment Analysis}
Aspect-based Sentiment Analysis (ABSA), as a popular natural language understanding task, has been extensively studied in the last decade~\cite{liu2012survey,schouten2015survey}. To tackle the ABSA challenge, an amount of neural network-based models were proposed, which can be roughly classified into three categories: LSTM-based methods~\cite{tang2016effective,wang2016attention,ma2017interactive,ma2018targeted,zhang2020knowledge}, CNN-based methods~\cite{xue2018aspect,li2018transformation,fan2018convolution,huang2019parameterized,chen2020inducing}, and Syntax-based methods~\cite{tang2020dependency,wang2020relational,hou-etal-2021-graph,li-etal-2021-dual-graph,pang-etal-2021-dynamic,zhong2022knowledge,lin2021deep}. 
The representative LSTM-based method is the target-dependent LSTM (TD-LSTM)~\cite{tang2016effective} that aims to take advantage of LSTM's ability to learn sequential patterns, and uses the LSTMs to extract aspect-specific features. Later, considering the complexity and inefficiency of LSTMs, some studies attempt to employ more efficient CNNs to capture the compositional structure and n-gram features. Xue and Li~\cite{xue2018aspect} introduce a gated convolution network to extract the contextual features and design the gate mechanism to output the useful sentiment features. For further enriching the sentiment information, numerous of studies involve explicitly leveraging the syntactic structure to establish the connection of aspect and opinion words. Zhang~\textit{et al.}~\cite{zhang-etal-2019-aspect} are the first to use the dependency trees to represent the sentence, and then leverage the graph convolution networks (GCNs) to exploit the syntactical information from the dependency trees.

In recent years, with the advancements of PLMs, a large amount of PLM-based ABSA models have emerged~\cite{he2019interactive,luo2019doer,xu2018double,he2018exploiting,chen2020enhancing,zhao2021knowledge}, which involve designing the network structure or injecting external knowledge in different ways. These methods have achieved promising performance on several widely-used ABSA benchmarks~\cite{pontiki-etal-2014-semeval,pontiki-etal-2015-semeval,jiang2019challenge}. However, most of them highly rely on numerous labeled data, which is expensive to obtain in some scenarios~\cite{yu2023cross}.

\subsection{Data Generation for ABSA}
To alleviate the above issue, a common approach is Data Generation (DG), which enlarges the training dataset by changing the original data or generating new data through various methods~\cite{mixup,wei-zou-2019-eda,kobayashi_contextual,wu2019conditional,Anaby2020do,wang2022promda}. In the context of ABSA, numerous DG methods have been proposed~\cite{wang-etal-2022-contrastive,yu2023cross,li2023data,chen2022unsupervised,hsu2021semantics,li2022generative,zhang2023target,wu2024novel}. For example, Wang~\textit{et~al.}~\cite{wang-etal-2022-contrastive} first train an in-domain generator and then leverage the generator to construct more multi-aspect samples. Wu~\textit{et~al.}~\cite{wu2024novel} propose a counterfactual data augmentation method to generate opinion expressions with reversed sentiment polarity.

Despite obtaining promising results, these DG methods still struggle to achieve optimal results in real-world scenarios. Since most of them attempt to augment the data by simply modifying the sentence structure or using pretrained models for text infilling, they have some shortcomings~\cite{yu2023cross}, \textit{e.g.}, poor fluency, and lack of diversity. Moreover, current DG methods usually rely on some existing labeled data, and might not be able to expand to real-world scenarios, in which the labeled data is unavailable. To this end, we propose a new zero-shot DG method, which is more effective and applicable, for alleviating the issue of data scarcity in ABSA.

\subsection{Large Language Models}
Recently, we have witnessed the great success of large language models (LLMs)~\cite{ouyang2022training,touvron2023llamav2,anil2023palm,openai2023gpt4,bai2023qwen} in many downstream language generation and understanding tasks. Owing to the instruction-tuning approach~\cite{wei2021finetuned}, LLMs can generate fluent and high-quality contexts following the human's instruction. Unfortunately, in the context of ABSA, directly using LLMs is not an optimal choice. Prior empirical studies~\cite{zhong2023can,han2023information} show that LLMs might under-perform the traditional BERT~\cite{devlin2018bert} models in some fine-grained language understanding tasks, \textit{e.g.}, ABSA. Thus, employing BERT-style PLMs is still a viable option for ABSA. Alternatively, in this paper, we attempt to take advantage of LLMs' abilities and enforce them to generate more high-quality data for boosting the performance of existing ABSA models. Notably, with the advancement of LLMs, some prior studies also attempt to prompt LLMs for data generation~\cite{ding2024data,bayer2023data}. However, in the field of ABSA, directly prompting LLMs struggles to generate the desired pseudo-label data, as LLMs are prone to hallucinations, leading to undesired data generation. To this end, we design a novel iterative data generation mechanism and a self-reflection data filtering module to better guide the data generation of LLMs.

\section{Methodology}
\label{sec:method}
In this section, we first briefly review the ABSA task and then present the details of our IDG, which contains three-stage processes: \ding{182} \textit{Aspect Extraction and Extension}, \ding{183} \textit{Pseudo Data Generation} and \ding{184} \textit{Evaluating and Filtering}. The framework of IDG is illustrated in Fig.~\ref{fig:IDG}.

\subsection{Problem Formulation} 
Given a sentence-aspect pair $\{S, T\}$, the goal of ABSA is to predict the sentiment polarity $y \in \{0,1,2\}$ of the sentence $S$ towards the aspect $T$, where 0, 1, and 2 denote the \textit{positive}, \textit{neutral} and \textit{negative} polarities, respectively. Note that $T$ is the subsequence of $S$. As mentioned in Sec.\ref{sec:intro}, there are usually limited labeled sentence-aspect pairs. Thus, we aim to generate the synthetic dataset $\mathcal{G}=\{(S_i, T_i)|i>i\}$ from an unsupervised text corpus $U=\{S_1,S_2,S_3,...,S_n\}$ with $n$ sentences.

\subsection{Iterative Data Generation}

\paragraph{\textbf{Aspect Extraction and Extension}}
Starting from an unsupervised corpus $U$, we first attempt to extract the aspects relevant to a specific domain. Specifically, we carefully design an aspect extraction (denoted as ``EX'') prompt to enforce the LLM to automatically extract domain-related aspects for each sentence $S_i \in U$. After doing that, we deduplicate the aspects and obtain the initial aspect set $A$. 

\begin{figure}[ht]
    \centering   
    \includegraphics[width=0.48\textwidth]{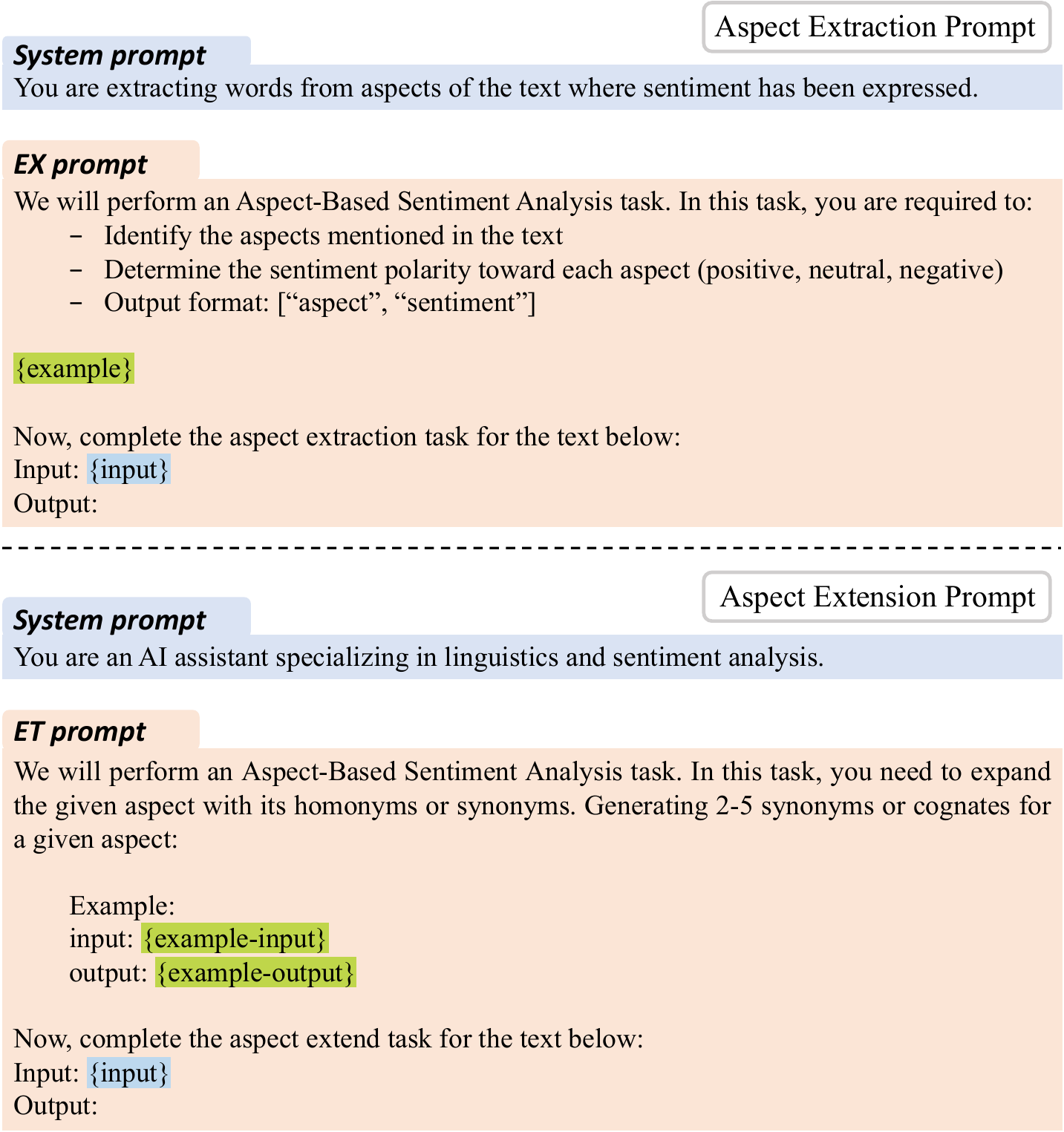}
    \caption{\textbf{Detailed prompts for aspect extraction and extension}.}
    \label{fig:ex_et}
\end{figure}

By analyzing the initial $A$, we find some aspects extracted by LLMs from unsupervised text data are prepositions (\textit{e.g.}, ``for/in"), conjunctions (\textit{e.g.}, ``and/or"), and similar words. These aspects do not carry actual meaning and cannot be used as aspects to generate accurate samples. Therefore, we choose to remove these noisy aspects. In practice, considering that aspects are generally nouns and their variants, we perform the part-of-speech processing with a Python library Textblob
on all candidate aspects of $A$ to remove those non-noun aspects. Then, to further improve the diversity of extracted aspects, we introduce an aspect extension module to expand $A$. In particular, for the Noun aspects in $A$, we enforce the LLM to expand them with their homonyms and synonyms by an aspect extension (denoted as ``ET'') prompt, as illustrated in Fig.~\ref{fig:ex_et}.
Lastly, the extend aspect set is merged into $A$. 

Moreover, since some extracted aspects inherently carry an emotional bias (such as ``loud noises'', which have a negative sentiment), randomly pairing these aspects with other sentiment polarities in a tuple would easily result in an inaccurate generation. Therefore, it is important to categorize aspects by different sentiment polarities before generating samples. By doing so, we can achieve more accurate fine-grained ABSA data generation. Specifically, we split the $A$ into three sub-sets with different sentiment polarities, \textit{i.e.}, positive aspects $A_{pos}$, negative aspects $A_{neg}$, neutral aspects $A_{neu}$, by performing a word sentiment analysis on each aspect.

\paragraph{\textbf{Pseudo Data Generation}} After obtaining the domain-related aspects, we then generate the pseudo labeled data, \textit{i.e.}, triplet $\{S_i, T_i, y_i\}$. Specifically, for each aspect sub-set, we append the aspects with their corresponding sentiment polarities to construct the aspect-sentiment set. For instance, for the aspect in $A_{pos}$, we append it with the positive polarity. Consequently, we can basically design a prompt to guide the data generation of LLMs based on the aspect-sentiment set. However, during the preliminary experiments, we found that as the generation of LLMs continued, LLMs suffer from the problem of repetitive generation, \textit{i.e.}, the generated samples tend to be similar and low-diversity. Hence, we propose a more powerful Iteration Teaching Analysis Prompt (denoted as ``ITAT''), which randomly selects samples from each round of generating samples as feedback to guide the next-round generation. By doing so, ITAT can prompt the LLMs to generate more richer and diverse pseudo-triplet data. 

\begin{figure}[t]
    \centering   
    \includegraphics[width=0.48\textwidth]{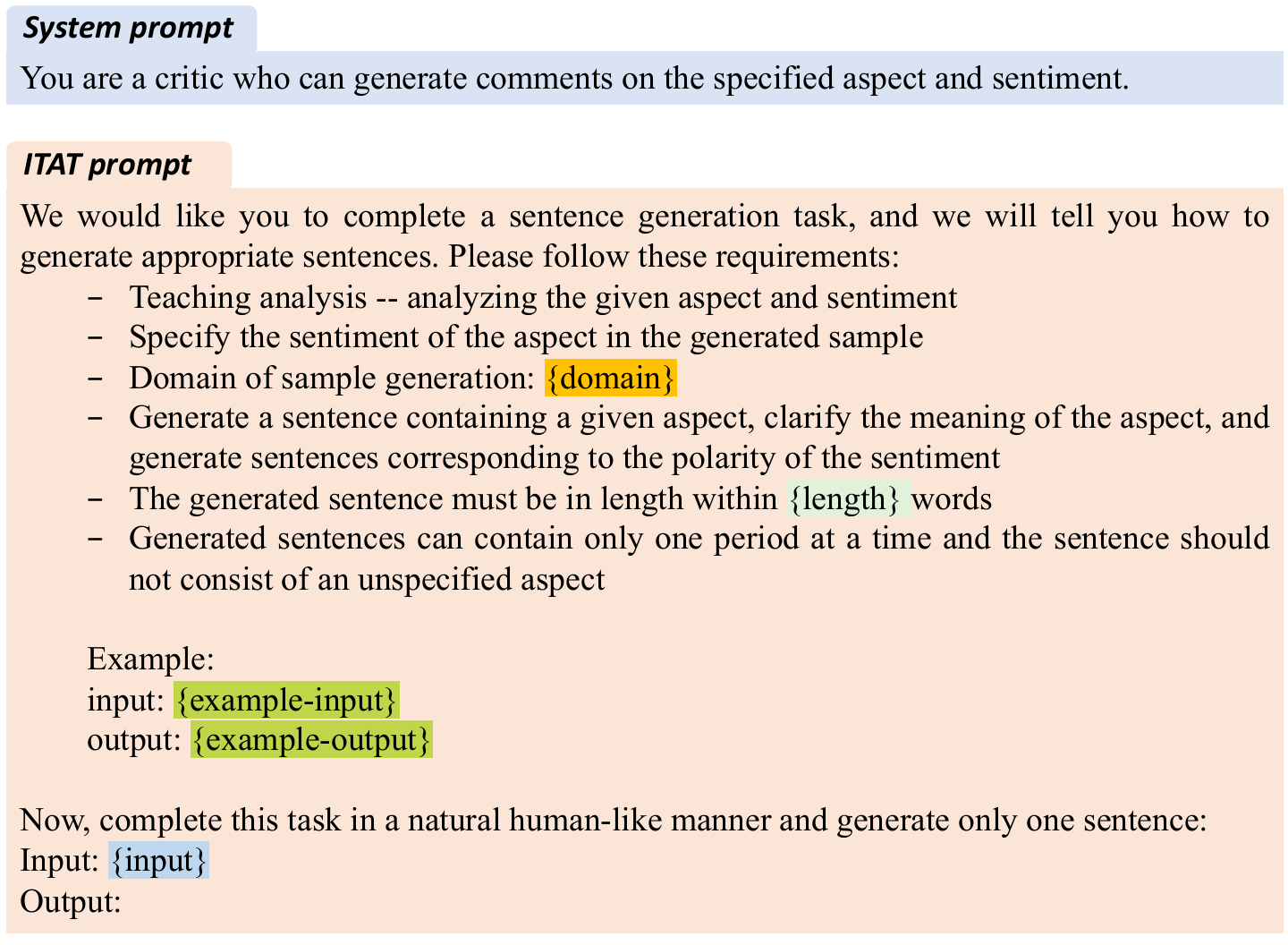}
    \caption{\textbf{Illustration of ITAT prompt}. The slots \{\texttt{example-input}\} and \{\texttt{example-output}\} denote the example of input-output pairs. The slots \{\texttt{domain}\} and \{\texttt{length}\} are the given sample domain and length. The slot \{\texttt{input}\} denotes the input aspect-sentiment pair.}
    \label{fig:itat_prompt}
\end{figure}

In particular, ITAT first designs some requirements to constrain the data generation of LLMs, and then iteratively leverages some high-quality previously-generated samples as demonstrations to further ensure the quality of current synthesis data. The detailed ITAT is shown in Fig.~\ref{fig:itat_prompt}.

\begin{figure}[t]
    \centering   
    \includegraphics[width=0.48\textwidth]{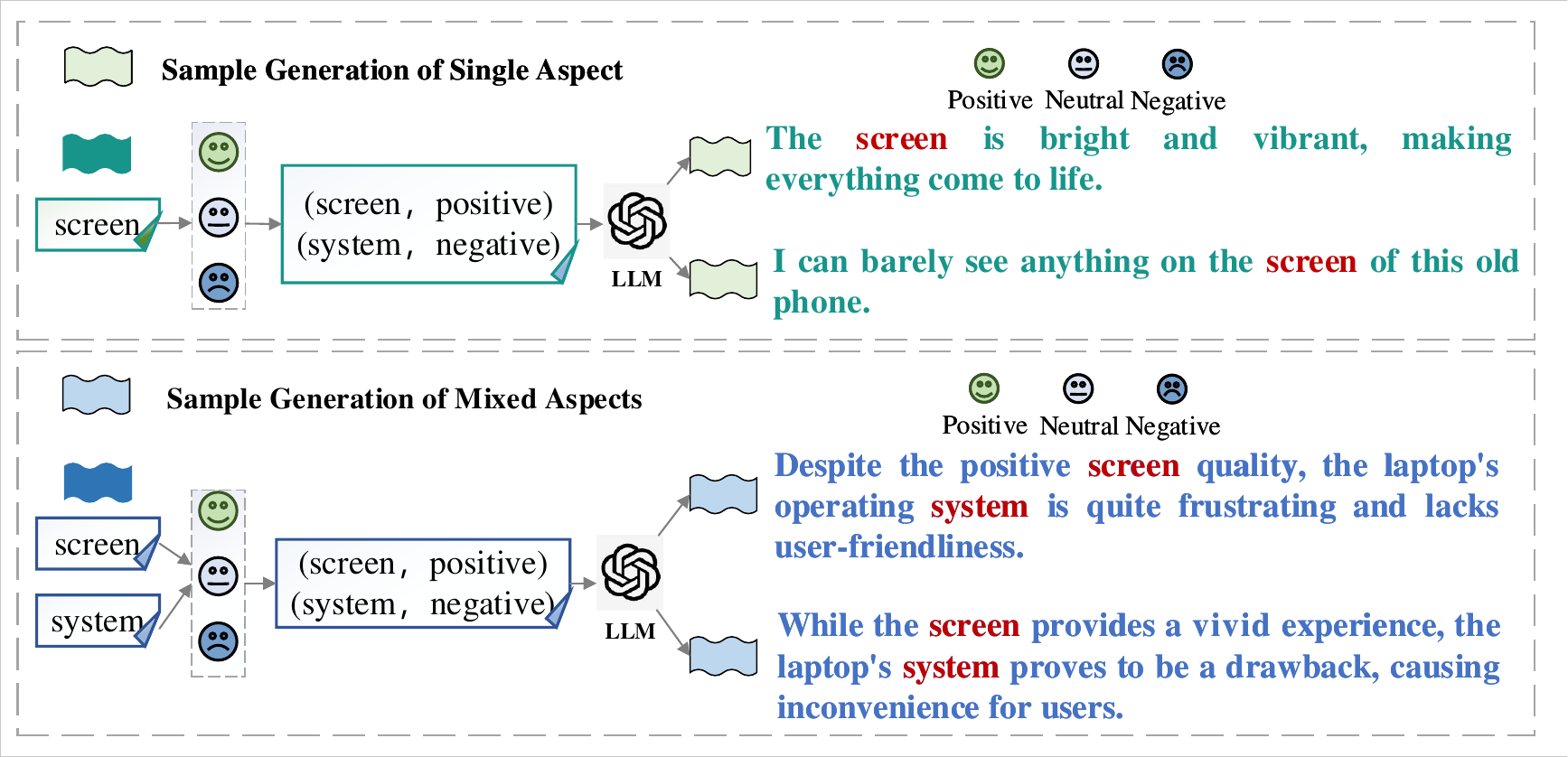}
    \caption{\textbf{Illustration of single-/multi-aspect data generation}. For ease of illustration, we only show some cases in the laptop domain.}
    \label{fig:DG_STR}
\end{figure}

Moreover, inspired by prior studies~\cite{wang-etal-2022-contrastive}, we recognize that multi-aspect data, \textit{i.e.}, data with multiple aspects in a sentence, is greatly beneficial to the training of ABSA models. To this end, in addition to the vanilla single-aspect pseudo data generation, we further utilize a multi-aspect pseudo data generation branch to obtain more complex yet effective multi-aspect data. To have a close look, we provide the illustrations of single- and multi-aspect pseudo data generation in Fig.~\ref{fig:DG_STR}.

\begin{figure*}
    \centering   
    \includegraphics[width=0.98\textwidth]{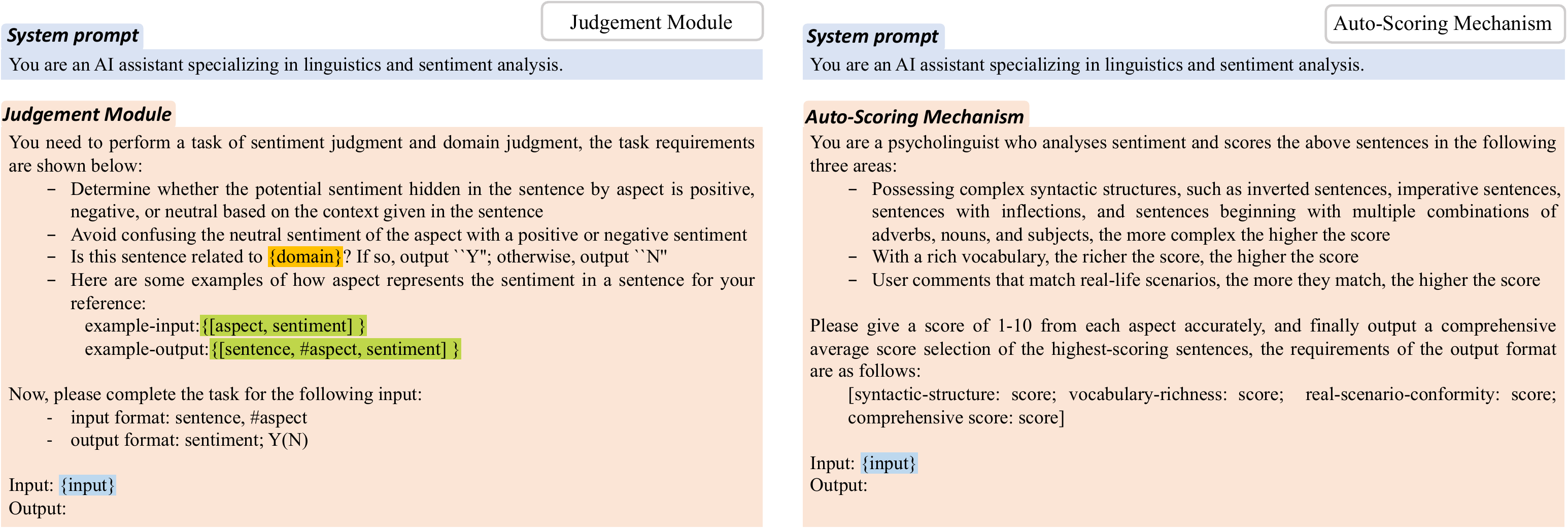}
    \caption{\textbf{Detailed prompts for discriminator}. The slots \{\texttt{domain}\} and \{\texttt{length}\} are the given sample domain and length. The slot \{\texttt{input}\} denotes the input sentence-aspect pair.}
    \label{fig:judgement_prompt}
\end{figure*}

\begin{figure}
    \centering   
    \includegraphics[width=0.42\textwidth]{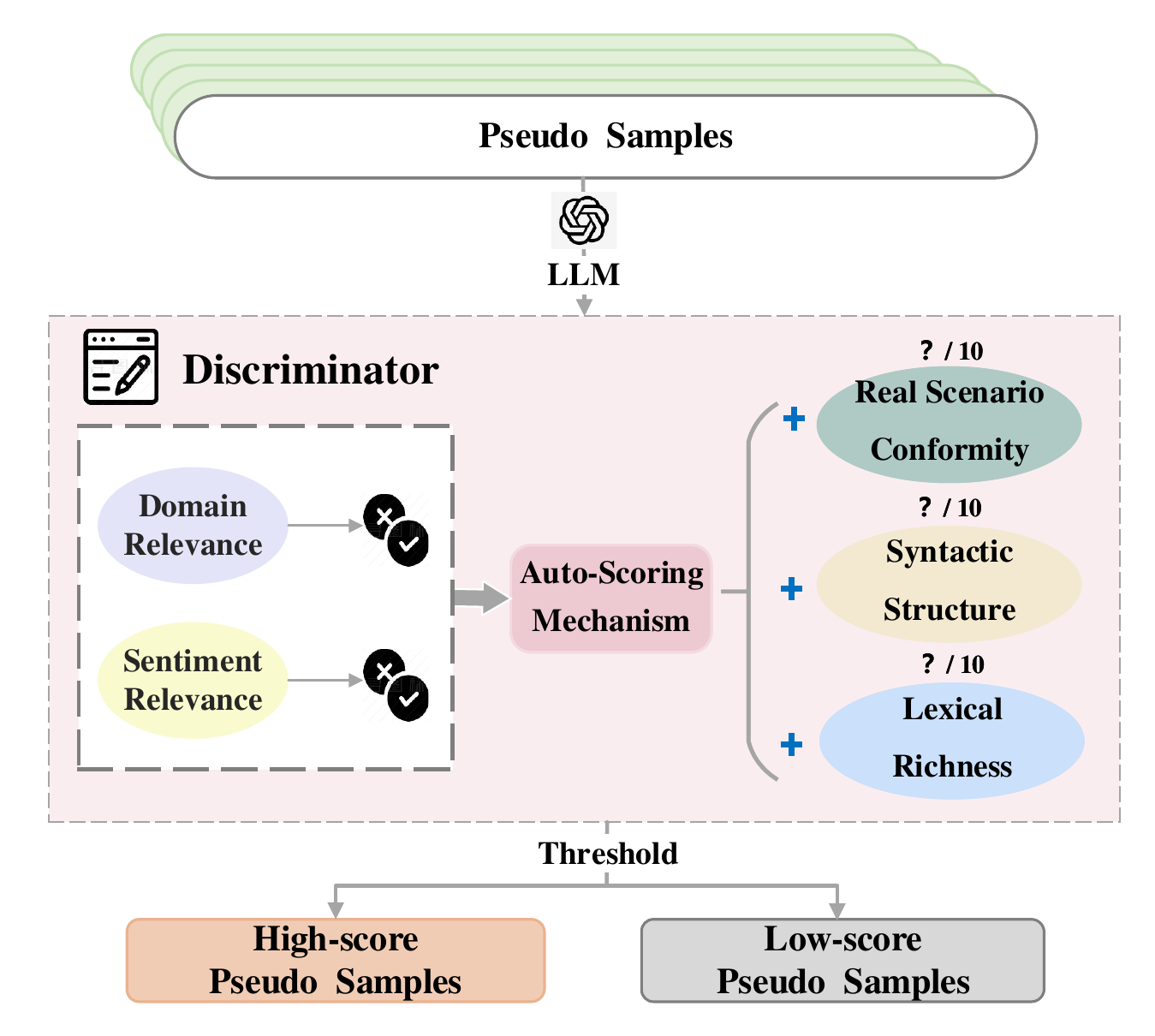}
    \caption{\textbf{Illustration of the discriminator.} Notably, ``Low-score Pseudo Sample'' denotes the low-quality data that will be filtered, while the ``High-score Pseudo Samples'' denotes the high-quality data that will be reused to guide the generation in stage \ding{183}.}
    \label{fig:disc}
\end{figure}

\paragraph{\textbf{Evaluating And Filtering}} Despite the powerful abilities of LLMs, they are prone to hallucinations and might unexpectedly generate low-quality data, hindering their performance. Thus, it is critical to evaluate the quality of generated data and filter the lower-quality ones. To achieve this goal, we introduce a new discriminator, as illustrated in Fig.~\ref{fig:disc}, containing a judgment module and an auto-scoring mechanism. Specifically, in the judgment module, we employ the popular \textbf{LLM-as-a-Judge} method to enforce the LLM to determine the domain relevance and sentiment relevance of synthesis data. That is, LLM is used to verify whether the synthesis data is relevant to the given domain and sentiment. 

After filtering the data with lower domain relevance and sentiment relevance, we further use the auto-scoring mechanism to quantitatively measure the data quality, in terms of Syntactic Structure, Lexical Richness, and Real Scenario Conformity. The scoring mechanism takes a sample judgment on a scale of 1-10, where larger scores mean higher data quality.
For filtering the low-quality data, we set a filtering threshold\footnote{The analysis of $\mathcal{T}$ can be found in Sec.\ref{sec:ablation}}$\mathcal{T}$. The data exceeding the threshold is used as final training data, while the others are discarded. Notably, for promoting the aforementioned ITAT strategies, we use the high-quality generated data as the feedback. By doing so, we can make full use of the self-reflection abilities of LLM to improve the generated data, ensuring both diversity and quality. The detailed prompts of the judgment module and auto-scoring mechanism are shown in Fig.~\ref{fig:judgement_prompt}.

\section{Experiments}
\label{sec:experiments}
\subsection{Experimental Setup}
\paragraph{\textbf{Task and Dataset}}
In this paper, we conduct main experiments on four public standard ABSA benchmarks, \textit{i.e.}, Laptop14, Restaurant14, Restaurant15, and Restaurant16. The Laptop14 and Restaurant14 datasets are from the SemEval2014 ABSA challenge \cite{pontiki-etal-2014-semeval}, and Restaurant15 and Restaurant16 are from the SemEval2015~\cite{pontiki-etal-2015-semeval} and SemEval2016~\cite{pontiki-etal-2016-semeval} challenges, respectively. Following prior studies~\cite{tang2019progressive,liu2023unified}, we remove a few instances with conflicting sentiment polarity.

To evaluate our IDG, we generate the synthetic data for each benchmark and compare the results training with the original data and generated data. Table~\ref{tab:dataset} shows the statistics of all used data in this work. Specifically, for the evaluation of aspects extracted by our IDG, we use ``Precision'' (P), ``Recall'' (R) and ``Macro-F1'' (F$_1$) as the metrics, while the ``Accuracy'' (Acc) and F$_1$ score are used to evaluate the final ABSA models. 

\begin{table}[t]
\centering
\caption{\textbf{Statistics of all used benchmarks}. Notably, ``Original'' denotes the original training and test sets of the benchmark, and ``Generated data'' denotes the synthetic data generated by our IDG. ``Rest14'', ``Rest15'' and ``Rest16'' refer to the Restaurant14, Restaurant15 and Restaurant16.}
\resizebox{0.95\columnwidth}{!}{%
\begin{tabular}{cccccccccc}
\toprule
\multirow{2}{*}{\textbf{Dataset}} &\multirow{2}{*}{\textbf{Type}} & \multicolumn{2}{c}{\textbf{Positive}} & \multicolumn{2}{c}{\textbf{Neutral}} & \multicolumn{2}{c}{\textbf{Negative}} \\
\cmidrule(r){3-4} \cmidrule(r){5-6} \cmidrule(r){7-8}
& &  \textbf{Train}      &  \textbf{Test}
&  \textbf{Train}      &  \textbf{Test}
&  \textbf{Train}      &  \textbf{Test}   \\
\midrule \midrule
\multirow{2}{*}{Laptop14} &Original &994&341&464&169&870&128\\
& Generated &1,051 &- &358 &- &919 &- \\ \midrule
\multirow{2}{*}{Rest14} &Original &2,164&728&637&196&807&196\\
& Generated &2,377 &- &548 &- &1,291 &- \\ \midrule
\multirow{2}{*}{Rest15} &Original &912&326&36&34&256&182\\
& Generated &1,572 &- &405 &- &1,631 &- \\ \midrule
\multirow{2}{*}{Rest16} &Original &1,240&469&69&30&439&117\\
& Generated &2,215 &- &184 &- &1,209 &- \\
\bottomrule
\end{tabular}
}
\label{tab:dataset}
\end{table}

\begin{table*}[ht]
\renewcommand\arraystretch{1.35}
\centering
\caption{
\textbf{Results of our IDG method on various baseline ABSA models}. Notably, ``Original data'' and ``Generated data'' denote that we train the models on the original ground-truth training data and our generated data, respectively. ``Mixed data'' means that we train on the mix of original and generated training data. Performance gains against the ``Original data'' are marked in \textcolor[RGB]{0,176,80}{green}, while the performance drops are marked in \textcolor[RGB]{205,92,92}{red}.
}
\setlength{\tabcolsep}{2mm}
\resizebox{\textwidth}{!}{%
\begin{tabular}{cccccccccc}
\toprule
\multicolumn{1}{c}{\multirow{2}{*}{\textbf{Model}}} & \multirow{2}{*}{\textbf{Dataset}} & \multicolumn{2}{c}{\textbf{Laptop14}} & \multicolumn{2}{c}{\textbf{Restaurant14}} & \multicolumn{2}{c}{\textbf{Restaurant15}} & \multicolumn{2}{c}{\textbf{Restaurant16}} \\ \cline{3-10} 
\multicolumn{2}{c}{} & \textbf{Acc} & \textbf{F$_1$} & \textbf{Acc} & \textbf{F$_1$} & \textbf{Acc} & \textbf{F$_1$} & \textbf{Acc} & \textbf{F$_1$} \\ \midrule \midrule
\multirow{3}{*}{\textbf{ATAE-LSTM}} & Original data & 79.50 & 75.50 & 83.42 & 75.03 & 83.39 & 68.59 & 91.41 & 77.08 \\
 & Generated data & 79.22\textcolor[RGB]{205,92,92}{$_{\downarrow0.29}$} & 75.64\textcolor[RGB]{0,176,80}{$_{\uparrow0.14}$} & 80.36\textcolor[RGB]{205,92,92}{$_{\downarrow3.06}$} & 70.52\textcolor[RGB]{205,92,92}{$_{\downarrow4.51}$} & 83.27\textcolor[RGB]{205,92,92}{$_{\downarrow0.12}$} & 70.42\textcolor[RGB]{0,176,80}{$_{\uparrow1.83}$} & 89.22\textcolor[RGB]{205,92,92}{$_{\downarrow2.19}$} & 76.89\textcolor[RGB]{205,92,92}{$_{\downarrow0.19}$} \\
 & Mixed data & \textbf{80.94}\textcolor[RGB]{0,176,80}{$_{\uparrow1.44}$} & \textbf{77.54}\textcolor[RGB]{0,176,80}{$_{\uparrow2.04}$} & \textbf{84.91}\textcolor[RGB]{0,176,80}{$_{\uparrow1.49}$} & \textbf{77.88}\textcolor[RGB]{0,176,80}{$_{\uparrow2.85}$} & \textbf{84.01}\textcolor[RGB]{0,176,80}{$_{\uparrow0.74}$} & \textbf{71.43\textcolor[RGB]{0,176,80}{$_{\uparrow2.84}$}} & \textbf{91.67}\textcolor[RGB]{0,176,80}{$_{\uparrow0.26}$} & \textbf{79.25}\textcolor[RGB]{0,176,80}{$_{\uparrow2.17}$} \\ \hline
\multirow{3}{*}{\textbf{ASGCN}} & Original data & 80.94 & 77.80 & 86.37 & 80.13 & 85.04 & 70.75 & 92.22 & 78.42 \\
 & Generated data & 80.62\textcolor[RGB]{205,92,92}{$_{\downarrow0.32}$} & 77.71\textcolor[RGB]{205,92,92}{$_{\downarrow0.09}$} & 82.95\textcolor[RGB]{205,92,92}{$_{\downarrow3.42}$} & 74.61\textcolor[RGB]{205,92,92}{$_{\downarrow5.52}$} & 85.48\textcolor[RGB]{0,176,80}{$_{\uparrow0.44}$} & 72.47\textcolor[RGB]{0,176,80}{$_{\uparrow1.72}$} & 89.74\textcolor[RGB]{205,92,92}{$_{\downarrow2.48}$} & 77.27\textcolor[RGB]{205,92,92}{$_{\downarrow1.15}$} \\
 & Mixed data & \textbf{82.03}\textcolor[RGB]{0,176,80}{$_{\uparrow1.09}$} & \textbf{79.17}\textcolor[RGB]{0,176,80}{$_{\uparrow1.37}$} & \textbf{87.23}\textcolor[RGB]{0,176,80}{$_{\uparrow0.86}$} & \textbf{81.45}\textcolor[RGB]{0,176,80}{$_{\uparrow1.32}$} & \textbf{86.21}\textcolor[RGB]{0,176,80}{$_{\uparrow1.17}$} & \textbf{74.55}\textcolor[RGB]{0,176,80}{$_{\uparrow3.80}$} & \textbf{93.11}\textcolor[RGB]{0,176,80}{$_{\uparrow0.89}$} & \textbf{82.43}\textcolor[RGB]{0,176,80}{$_{\uparrow4.01}$} \\ \hline
\multirow{3}{*}{\textbf{BERT-SPC}} & Original data & 78.68 & 74.82 & 84.82 & 78.08 & 83.95 & 69.91 & 90.42 & 76.61 \\
 & Generated data & 77.02\textcolor[RGB]{205,92,92}{$_{\downarrow1.66}$} & 73.97\textcolor[RGB]{205,92,92}{$_{\downarrow0.85}$} & 85.24\textcolor[RGB]{0,176,80}{$_{\uparrow0.42}$} & 72.34\textcolor[RGB]{205,92,92}{$_{\downarrow5.74}$} & 83.39\textcolor[RGB]{205,92,92}{$_{\downarrow0.57}$} & 69.70\textcolor[RGB]{205,92,92}{$_{\downarrow0.21}$} & 88.66\textcolor[RGB]{205,92,92}{$_{\downarrow1.76}$} & 72.75\textcolor[RGB]{205,92,92}{$_{\downarrow3.86}$} \\
 & Mixed data & \textbf{80.09}\textcolor[RGB]{0,176,80}{$_{\uparrow1.41}$} & \textbf{77.13}\textcolor[RGB]{0,176,80}{$_{\uparrow2.31}$} & \textbf{85.62}\textcolor[RGB]{0,176,80}{$_{\uparrow0.80}$} & \textbf{78.45}\textcolor[RGB]{0,176,80}{$_{\uparrow0.37}$} & \textbf{85.24}\textcolor[RGB]{0,176,80}{$_{\uparrow1.29}$} & \textbf{70.65}\textcolor[RGB]{0,176,80}{$_{\uparrow0.74}$} & \textbf{90.75}\textcolor[RGB]{0,176,80}{$_{\uparrow0.33}$} & \textbf{77.37}\textcolor[RGB]{0,176,80}{$_{\uparrow0.76}$} \\ \hline
\multirow{3}{*}{\textbf{R-GAT}} & Original data & 78.37 & 73.92 & 86.34 & 80.74 & 83.58 & 71.48 & 91.72 & 77.77 \\
 & Generated data & 78.58\textcolor[RGB]{0,176,80}{$_{\uparrow0.21}$} & 75.67\textcolor[RGB]{0,176,80}{$_{\uparrow1.75}$} & 81.79\textcolor[RGB]{205,92,92}{$_{\downarrow4.55}$} & 74.60\textcolor[RGB]{205,92,92}{$_{\downarrow6.14}$} & 84.32\textcolor[RGB]{0,176,80}{$_{\uparrow0.74}$} & 69.14\textcolor[RGB]{205,92,92}{$_{\downarrow2.34}$} & 88.96\textcolor[RGB]{205,92,92}{$_{\downarrow2.76}$} & 75.64\textcolor[RGB]{205,92,92}{$_{\downarrow2.13}$} \\
 & Mixed data & \textbf{80.56}\textcolor[RGB]{0,176,80}{$_{\uparrow2.19}$} & \textbf{77.08}\textcolor[RGB]{0,176,80}{$_{\uparrow3.16}$} & \textbf{87.50}\textcolor[RGB]{0,176,80}{$_{\uparrow1.16}$} & \textbf{82.04}\textcolor[RGB]{0,176,80}{$_{\uparrow1.30}$} & \textbf{85.06}\textcolor[RGB]{0,176,80}{$_{\uparrow1.48}$} & \textbf{73.36}\textcolor[RGB]{0,176,80}{$_{\uparrow2.28}$} & \textbf{92.05}\textcolor[RGB]{0,176,80}{$_{\uparrow0.33}$} & \textbf{78.80}\textcolor[RGB]{0,176,80}{$_{\uparrow1.03}$} \\ \hline
\multirow{3}{*}{\textbf{KGAN}} & Original data & 82.34 & 79.17 & 86.55 & 81.47 & 86.40 & 73.89 & 92.81 & 81.17 \\
 & Generated data & 80.47\textcolor[RGB]{205,92,92}{$_{\downarrow1.87}$} & 76.83\textcolor[RGB]{205,92,92}{$_{\downarrow2.34}$} & 81.70\textcolor[RGB]{205,92,92}{$_{\downarrow4.85}$} & 74.11\textcolor[RGB]{205,92,92}{$_{\downarrow0.19}$} & 85.11\textcolor[RGB]{205,92,92}{$_{\downarrow7.36}$} & 72.11\textcolor[RGB]{205,92,92}{$_{\downarrow1.29}$} & 89.22\textcolor[RGB]{205,92,92}{$_{\downarrow3.59}$} & 77.71\textcolor[RGB]{205,92,92}{$_{\downarrow3.46}$} \\
 & Mixed data & \textbf{82.49}\textcolor[RGB]{0,176,80}{$_{\uparrow0.15}$} & \textbf{79.62}\textcolor[RGB]{0,176,80}{$_{\uparrow0.45}$} & \textbf{87.50}\textcolor[RGB]{0,176,80}{$_{\uparrow0.95}$} & \textbf{81.86}\textcolor[RGB]{0,176,80}{$_{\uparrow0.39}$} & \textbf{87.13}\textcolor[RGB]{0,176,80}{$_{\uparrow0.73}$} & \textbf{75.17}\textcolor[RGB]{0,176,80}{$_{\uparrow1.28}$} & \textbf{92.95}\textcolor[RGB]{0,176,80}{$_{\uparrow0.14}$} & \textbf{82.83}\textcolor[RGB]{0,176,80}{$_{\uparrow1.66}$} \\ \bottomrule
\end{tabular}
}
\label{tab:main_res}
\end{table*}

\paragraph{\textbf{Implementation}}
For simulating the real-world scenarios, we use the unlabeled sentences in training sets of the above ABSA benchmarks (\textit{i.e.}, ignoring the aspect and polarity information) as the initial unsupervised corpus for our IDG. The aspects of the original training sets are used as gold labels to evaluate our extracted aspects. After obtaining the synthesis ABSA data, we train the models with these data and evaluate them on the test sets of the above benchmarks. Specifically, we use the powerful GPT-3.5-turbo
\footnote{\url{https://platform.openai.com/docs/models/gpt-3-5-turbo}}
as the LLM in our IDG. The filtering threshold $\mathcal{T}$ used in IDG is set as 6. For each benchmark, we enforce IDG to generate the ABSA data, the number of which is similar to that of the original training set. 

\paragraph{\textbf{Baseline Models}}
To investigate the effectiveness of our IDG, we mainly apply it to improve five representative baseline ABSA models,
including:
\begin{itemize}
    \item \textbf{ATAE-LSTM}~\cite{wang-etal-2016-attention}: A LSTM-based model for ABSA using aspect embedding and attention mechanism.
    \item \textbf{ASGCN}~\cite{zhang-etal-2019-aspect}: It is the first ABSA model to represent sentences with dependency trees and use GCN to explore the syntactical information.
    \item \textbf{BERT-SPC}~\cite{song2019attentional}: BERT-SPC feeds sequence ``[CLS] + context + [SEP] + target + [SEP]'' into the basic BERT model for sentence pair classification task.
    \item \textbf{R-GAT}~\cite{wang-etal-2020-relational}: It uses a novel aspect-oriented dependency tree structure to reshape and prune ordinary dependency parse trees to better model syntax information.
   \item \textbf{KGAN} \cite{Zhong2022KnowledgeGA}: A novel knowledge graph augmented network encodes different types of information as multiview representations to enrich the semantic features.
\end{itemize}
For each model, we utilize the BERT-base-uncased
\footnote{\url{https://huggingface.co/google-bert/bert-base-uncased}}
as the backbone and train it following the default settings in the original papers. 

\paragraph{\textbf{Compared Methods}}
We conduct the main results in 3 different settings, \textit{i.e.}, 1) ``Original data'': training the ABSA models with the original labeled ABSA data, 2) ``Generated data'': training with only the synthetic data generated by our IDG and 3) ``Mixed data'': training with the mix of original data and our generated data. We additionally compare IDG with several representative data generation methods, including:
\begin{itemize}
     \item \textbf{Back Translation (BT)}~\cite{sennrich-etal-2016-improving}: It is a sentence-level data augmentation method, which first translates a sentence to another language and then translates it back to the original language.
    \item \textbf{EDA}~\cite{wei-zou-2019-eda}: It is a simple word-level data augmentation technique containing four operations: synonym substitution, random insertion, random exchange, and random deletion.
    \item \textbf{CBERT}~\cite{wu2019conditional}:         It integrates label information into the masked language modeling task to realize the prediction of replacement words, considering not only context but also label information
    \item \textbf{C3DA}~\cite{wang-etal-2022-contrastive}: It uses a pre-trained generator to construct the synthetic multi-aspect training dataset.
    \item \textbf{LLM-Rewriting}: Given the labeled ABSA corpus, it uses the LLM to rewrite existing samples for augmenting training data.
    \item \textbf{LLM-Annotating}: Similar to our IDG, it starts from an unlabeled sentence corpus and directly enforces the LLM to 1) extract the aspects and 2)  generate pseudo-label ABSA data with in-context learning.
\end{itemize}
Notably, \textbf{BT}, \textbf{EDA}, \textbf{CBERT} and \textbf{C3DA} are the traditional data generation methods that rely on the existing labeled training samples. Conversely, \textbf{LLM-Rewriting}, \textbf{LLM-Annotating} and our \textbf{IDG} are zero-shot LLM-based data generation methods that do not require the labeled data.

\begin{table}[t]
\centering
\caption{\textbf{Evaluation on aspects extracted by IDG with different strategies.} Notably,
``Zero-shot'' refers to the aspects extracted in a zero-shot manner, ``Few-shot$_{related}$'' refers to few-shot extraction using domain-related demonstrations, and ``Few-shot$_{random}$'' refers to the few-shot extraction using random demonstrations.}
\resizebox{0.95\columnwidth}{!}{%
\begin{tabular}{cccccc}
\toprule
\textbf{Method} & \bf Metric   & \textbf{Laptop14} & \textbf{Rest14} & \textbf{Rest15} & \textbf{Rest16} \\ \midrule 
\midrule
\multirow{3}{*}{Zero-Shot} & P & 36.04    & 44.24  & 44.38  & 40.2   \\ 
                           & R    & 69.27    & 65.65  & 72.82  & 65.04  \\ 
                           & F$_1$        & 47.41    & 52.86  & 55.15  & 49.69  \\ \hline
\multirow{3}{*}{Few-shot$_{related}$}  & P & 46.79    & 59.85  & 60.34  & 57.31  \\  
                           & R    & 73.12    & 70.04  & 72.82  & 73.19  \\ 
                           & F$_1$  & 57.07    & 64.55  & 65.99  & 64.28  \\ \hline
\multirow{3}{*}{Few-shot$_{random}$} & P & 45.72    & 48.00     & 50.25  & 46.36  \\ 
                           & R    & \textbf{79.77 }   & \textbf{79.84}  & \textbf{82.15}  & \textbf{80.30}   \\ 
                           & F$_1$        & 58.13    & 59.95  & 62.36  & 58.79  \\ 
\bottomrule
\end{tabular}
}
\label{tab:aspect_extract_res}
\end{table}

\subsection{Main Results}
\subsubsection{Evaluation on the Extracted Aspect}
In our IDG, the performance of final ABSA models highly relies on the relevance between extracted aspects and gold aspects. Here, to verify whether IDG can extract the relevant aspects, we evaluate the aspects extracted by different strategies (``Zero-shot'', ``Few-shot$_{related}$'' and ``Few-shot$_{random}$'') of IDG and report the contrastive results in Table~\ref{tab:aspect_extract_res}. Specifically, ``Zero-shot'' means that we directly enforce the LLM to extract the aspects in the zero-shot manner. ``Few-shot$_{related}$'' and ``Few-shot$_{random}$'' denote that we select some aspect-sentence pairs as demonstrations to guide the aspect extraction of LLM, where the former uses domain-related demonstrations and the later uses random demonstrations.

As seen, given some demonstrations, IDG can extract more relevant aspects, indicating the superiority of few-shot learning. Interestingly, compared to the domain-related demonstrations, IDG with random demonstrations performs better. We conjecture that domain-related demonstrations might be too similar and hinder the diversity of extracted aspects, thus leading to sub-optimal performance. Notably, ``Few-shot$_{random}$'' performs best, and we thus use it as the default setting in the following content.

\subsubsection{Evaluation on the Generated Data}
In this part, we perform the evaluation of the synthetic data generated by IDG. The contrastive results are presented in Table~\ref{tab:main_res} and~\ref{tab:comparable_res}, from which we observe that:

\textbf{Models trained on the generated data partially outperforms those trained on the ground-truth data.} As seen in Table~\ref{tab:main_res}, training with only the generated data achieves remarkable or even better performance than on the ground-truth data, \textit{e.g.}, +1.75\% F$_1$ score of R-GAT in the Laptop14, and +0.42\% accuracy of BERT-SPC in the Restaurant14. 

\begin{table}
\renewcommand\arraystretch{1.15}
\centering
\caption{\textbf{Comparison of different data generation methods.}}
\setlength{\tabcolsep}{12pt}
\resizebox{1\columnwidth}{!}{%
\begin{tabular}{lcccc}
\toprule
\multirow{2}{*}{\textbf{Method}} & \multicolumn{2}{c}{\textbf{Laptop14}} & \multicolumn{2}{c}{\textbf{Restaurant14}}  \\ \cline{2-5} 
                                  & \textbf{Acc}           & \textbf{F$_1$}           & \textbf{Acc}          & \textbf{F$_1$}            \\ \midrule \midrule
R-GAT                             & 78.37         & 73.92        & 86.34        & 80.74              \\ \hline
+BT~\cite{sennrich-etal-2016-improving}                 & 79.70          & 75.01        & 86.85        & 81.02              \\ 
+EDA~\cite{wei-zou-2019-eda}                              & 78.59         & 74.82        & 86.52        & 81.47              \\ 
+CBERT~\cite{wu2019conditional}                            & 78.62         & 74.96        & 87.01        & 82.19              \\ 
+C3DA~\cite{wang-etal-2022-contrastive}                             & 79.16         & 75.40         & 87.22        & \textbf{82.69}               \\ \hdashline
+LLM-Rewriting                            & 79.53        &75.38       & 83.11       & 74.35             \\ 
+LLM-Annotating                             & 79.38         & 75.39         & 82.99      & 75.58               \\
\rowcolor{gray!20} \bf +IDG (Ours)                  & \textbf{80.25}         & \textbf{76.18 }       & \textbf{87.50}         & 82.04               \\ \bottomrule
\end{tabular}
}
\label{tab:comparable_res}
\end{table}

Although training with synthesis data might under-outperform the manually labeled data in some cases, we should emphasize that manual annotation is time-consuming and costly, while our IDG is more efficient and cheap. In general, these results show that IDG has the potential to generate high-quality labeled ABSA data, similar to the manually annotated data.

\textbf{IDG brings consistent and significant performance gains among all baseline models and tasks.} By combining the ground-truth data with our generated data, we find that there are consistent and significant performance gains among all settings, up to +4.01\% F$_1$ score. These results show that our IDG can be used to effectively generate the domain-specific ABSA data and is beneficial to various baseline ABSA models. Thus, we believe that our IDG has the great potential to be applied in real-world scenarios. 

\begin{table}
\renewcommand\arraystretch{1.15}
\setlength{\tabcolsep}{13.5pt}
\centering
\caption{\textbf{Ablation study of aspect extension module in IDG}. ``-w/o Extension'' means that we do not extend the aspect set in IDG. Laptop14 is used for evaluation.
} 
\resizebox{0.95\columnwidth}{!}{%
\begin{tabular}{cccc}
\toprule
\textbf{Model} & \textbf{Method} &  \textbf{Acc} & \textbf{F$_1$} \\ \midrule \midrule
\multirow{3}{*}{ASGCN} & IDG (Ours) & \bf 80.62 & \bf 77.71 \\
 & -w/o Extension & 80.15 & 77.50 \\ 
 & $\Delta (\downarrow)$ &\textcolor[RGB]{205,92,92}{\textbf{$\downarrow$ 0.47}}  & \textcolor[RGB]{205,92,92}{\textbf{$\downarrow$ 0.21}} \\
 \midrule
\multirow{3}{*}{R-GAT} & IDG (Ours) & \bf 78.58 & \bf 75.67 \\
 & -w/o Extension & 78.27 & 74.77 \\
  &$\Delta (\downarrow)$ & \textcolor[RGB]{205,92,92}{\textbf{$\downarrow$ 0.31}} & \textcolor[RGB]{205,92,92}{\textbf{$\downarrow$ 0.90}} \\
 \bottomrule
\end{tabular}
}
\label{tab:aspect_ex_dg_res}
\end{table}

\textbf{IDG outperforms the other counterparts by a clear margin.} In Table~\ref{tab:comparable_res}, we compare our method with the other data generation counterparts on the R-GAT model. Considering that some data generation methods (\textit{i.e.}, BT, EDA, CBERT and C3DA) require some existing labeled data, we conduct experiments in the ``Mixed data'' setting. That is, we use several methods to generate the synthesis data and merge them with the original labeled data for training the ABSA model. From the results in Table~\ref{tab:comparable_res}, we observe that IDG performs better than the others in most settings without using labeled data. More specifically, the other LLM-based methods (\textit{i.e.}, LLM-Rewriting and LLM-Annotating) struggle to improve the ABSA performance. One of the reasons is that these methods generate the low-quality pseudo-label data, which disturbs the training of models. This indicates the necessary of iterative generation and low-quality data filtering.

\subsection{Ablation Study}
\label{sec:ablation}
We evaluate the impact of each component of our IDG, including 1) aspect extension module, 2) sample generation strategies, 3) discriminator for filtering the low-quality data, and 4) filtering threshold $\mathcal{T}$. Notably, in the following content, for better investigating the effectiveness of IDG, we uniformly conduct experiments in the ``Generated data'' setting. That is, we directly train the ABSA models using the synthesis data generated by IDG.

\begin{table}[t]
\renewcommand\arraystretch{1.35}
\centering
\caption{\textbf{Analysis of different generation strategies}. ``Single-aspect'' denotes that we only generate the samples with a single aspect in a sentence, and ``Multi-aspect'' means that there are multiple aspects in a generated sentence. Here, we report the results on the Laptop14 benchmark.
} 
\resizebox{0.95\columnwidth}{!}{%
\begin{tabular}{ccccc}
\toprule
\multirow{2}{*}{\textbf{Method}} & \multicolumn{2}{c}{\textbf{ASGCN}} & \multicolumn{2}{c}{\textbf{R-GAT}} \\ 
\cline{2-5}
 & \textbf{Acc} & \textbf{F$_1$} & \textbf{Acc} & \textbf{F$_1$} \\
\midrule \midrule
Single-aspect & 76.09 & 72.42 & 72.88 & 68.71 \\   
\quad +Multi-aspect & \textbf{80.62}\textcolor[RGB]{0,176,80}{$_{\uparrow4.53}$} & \textbf{77.71}\textcolor[RGB]{0,176,80}{$_{\uparrow5.29}$} & \textbf{78.58}\textcolor[RGB]{0,176,80}{$_{\uparrow5.70}$} & \textbf{75.67}\textcolor[RGB]{0,176,80}{$_{\uparrow6.96}$} \\ 
 \bottomrule
\end{tabular}
}
\label{tab:stra_res}
\end{table}

\paragraph{\textbf{Impact of aspect extension}} As mentioned in \S\ref{sec:method}, we expand the aspect set to improve its diversity. Here, to verify its effectiveness, we compare IDG with a simple alternative, ``-w/o Extension'', \textit{i.e.}, removing the aspect extension module. Taking the ASGCN and R-GAT as examples, we provide the contrastive results on Laptop14 benchmark in Table~\ref{tab:aspect_ex_dg_res}. It can be seen that removing the aspect extension causes clear performance degradation, indicating that more diverse aspects are beneficial to the final ABSA performance.

\begin{table}[t]
\renewcommand\arraystretch{1.15}
\setlength{\tabcolsep}{12pt}
\centering
\caption{\textbf{Ablation study of discriminator in IDG}. ``-w/o Discriminator'' means that we directly use the generated data without filtering as final training data. Here, we report the results on the Laptop14 benchmark.} 
\resizebox{0.95\columnwidth}{!}{%
\begin{tabular}{cccc}
\toprule
\bf \textbf{Model} & \bf \textbf{Method} & \bf \textbf{Acc} & \bf \textbf{F$_1$} \\ \midrule \midrule
\multirow{3}{*}{ATAE-LSTM} & IDG (Ours) & \bf 79.22 & \bf 75.64 \\ 
 & -w/o Discriminator & 76.06 & 72.80 \\ 
 & $\Delta (\downarrow)$ &\textcolor[RGB]{205,92,92}{\textbf{$\downarrow$ 3.16}}  & \textcolor[RGB]{205,92,92}{\textbf{$\downarrow$ 2.84}} \\
 \midrule
\multirow{3}{*}{ASGCN} & IDG (Ours)  & \bf 80.62 & \bf 77.71 \\ 
 & -w/o Discriminator & 74.84 & 70.99 \\
 & $\Delta (\downarrow)$ &\textcolor[RGB]{205,92,92}{\textbf{$\downarrow$ 5.78}}  & \textcolor[RGB]{205,92,92}{\textbf{$\downarrow$ 6.72}} \\
 \midrule
\end{tabular}
}
\label{tab:discri_res}
\end{table}

\paragraph{\textbf{Impact of different sample generation strategies}} In the sample generation phase of IDG, we use two different strategies, \textit{i.e.}, single-aspect and multi-aspect generation. Specifically, the latter strategy is to simulate the multi-aspect problem~\cite{wang-etal-2022-contrastive} in ABSA. Notably, for a fair comparison, we generate the same number of training data for both strategies and present the compared results in Table~\ref{tab:stra_res}. As seen, by generating more multi-aspect data, IDG brings consistent and significant performance gains against the vanilla single-aspect data. This is similar to the findings~\cite{wang-etal-2022-contrastive}, as training on multi-aspect data can encourage the models to extract more fine-grained aspect-specific information, thus leading to better performance.


\begin{figure}
    \centering   
    \includegraphics[width=0.4\textwidth]{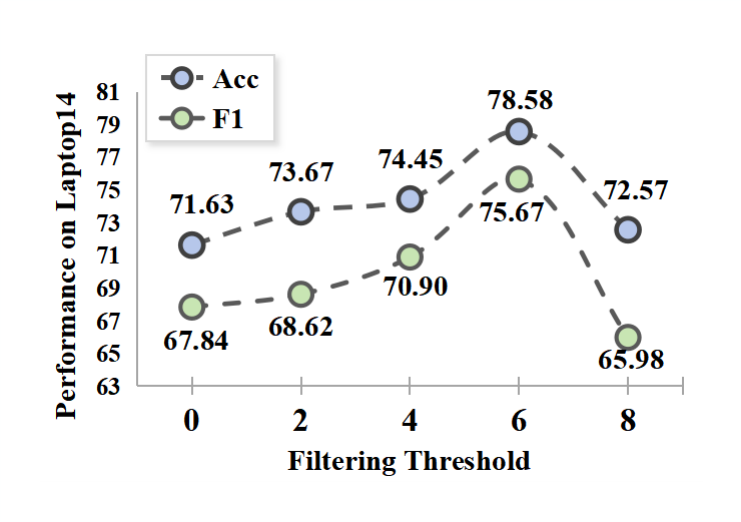}
    \caption{
    \textbf{Parameter analysis of filtering threshold $\mathcal{T}$}.  
    }
    \label{fig:ST}
\end{figure}

\paragraph{\textbf{Impact of discriminator}} In our IDG, we introduce a discriminator to filter the low-quality generated data. Here, we verify its effectiveness and report the contrastive results of ATAE-LSTM and ASGCN on Laptop14 in Table~\ref{tab:discri_res}. Compared to the full IDG method, removing the discriminator (\textit{i.e.}, directly using the generated data without filtering) will lead to much performance drops. This highlights the importance of filtering the low-quality data, and indicates that data quality is more important than the data quantity for the field of ABSA.

\paragraph{\textbf{Parameter analysis on $\mathcal{T}$}}
The threshold $\mathcal{T}$, which is used to control the threshold for filtering data, is an important hyper-parameter in IDG. Here, we analyze its influence by evaluating the performance with different $\mathcal{T}$, spanning \{0, 2, 4, 6, 8\}. Notably, for a fair comparison, we generate the same number of training data for each setting. Fig.~\ref{fig:ST} illustrates the contrastive results of R-GAT on Laptop14. With the increasing of $\mathcal{T}$ in a certain range (\textit{i.e.}, 0 to 6), IDG continues achieving better performance. This indicates that filtering low-quality data is beneficial. Conversely, too large $\mathcal{T}$ values (\textit{e.g.}, 8) lead to performance degradation, as filtering too much data might lead to limited available data for training. More specifically, $\mathcal{T} = 6$ performs best, thus leaving as the default setting.

\begin{figure}[t]
    \centering   
    \includegraphics[width=0.45\textwidth]{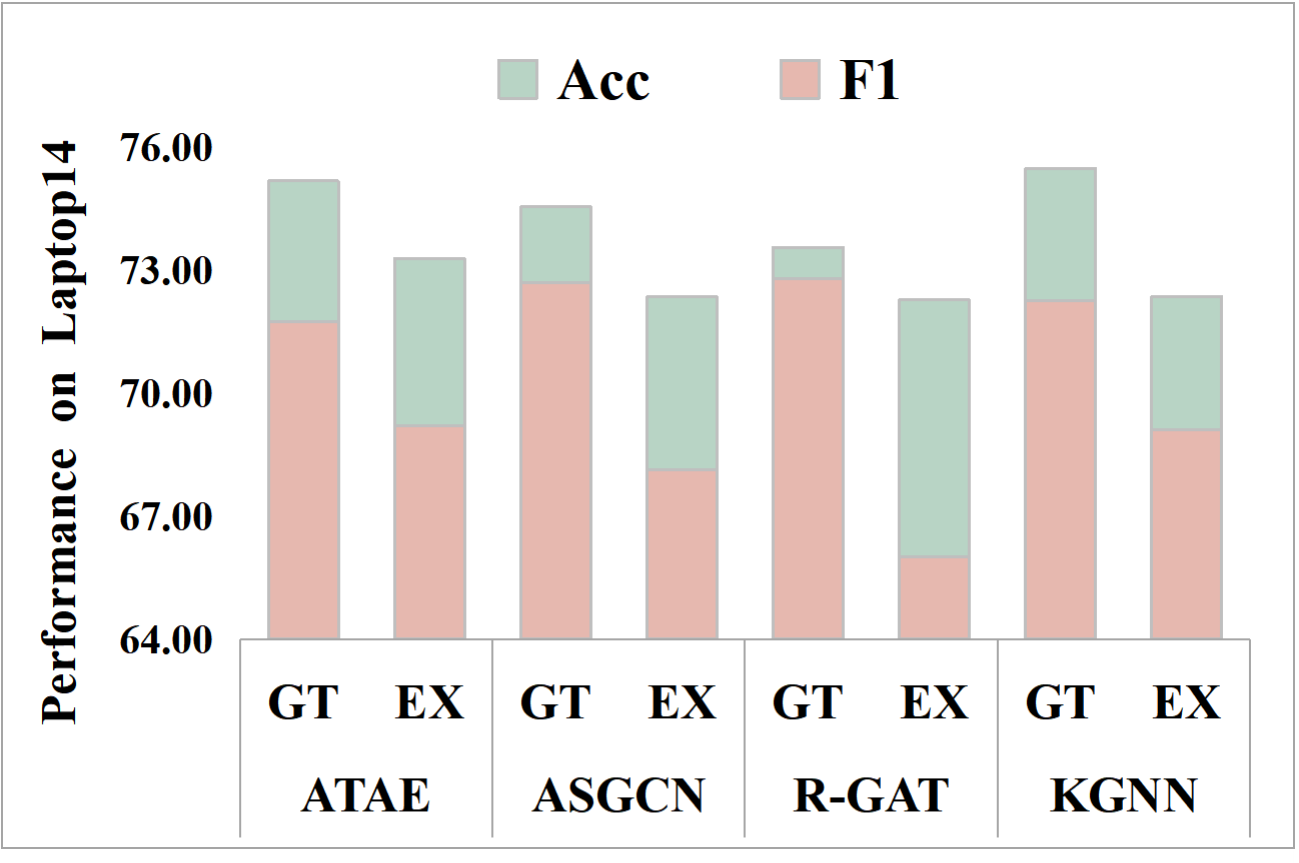}
    \caption{\textbf{Impact of accuracy of extracted aspects.} We replace the extracted aspects with gold ones in IDG and verify whether gold aspects can lead to better performance. ``GT'' and ``EX'' denote the gold and extracted aspects. 
    }
    \label{fig:ASPECT_IN}
\end{figure}

\subsection{Discussion and Analysis}

In this part, we perform more in-depth analyses to further explore the underlying mechanism of our proposed IDG, covering 1) the impact of the accuracy of extracted aspects, 
2) the effect of ITAT prompt, and 3) an analysis of the number of generated data.

\paragraph{\textbf{Impact of the accuracy of extracted aspects}}
Intuitively, based on more accurate aspects, IDG can generate more relevant training data and bring more performance gains. To verify it, we use the gold aspects in the original training sets as the upper bound to guide the generation of IDG. The contrastive results are illustrated in Fig.~\ref{fig:ASPECT_IN}, from which we find that IDG with gold aspects indeed achieves much better results. This indicates that \textit{the performance of IDG relies on the accuracy of extracted aspects and more accurate aspects can result in better performance}.

\begin{figure}[t]
    \centering   
    \includegraphics[width=0.43\textwidth]{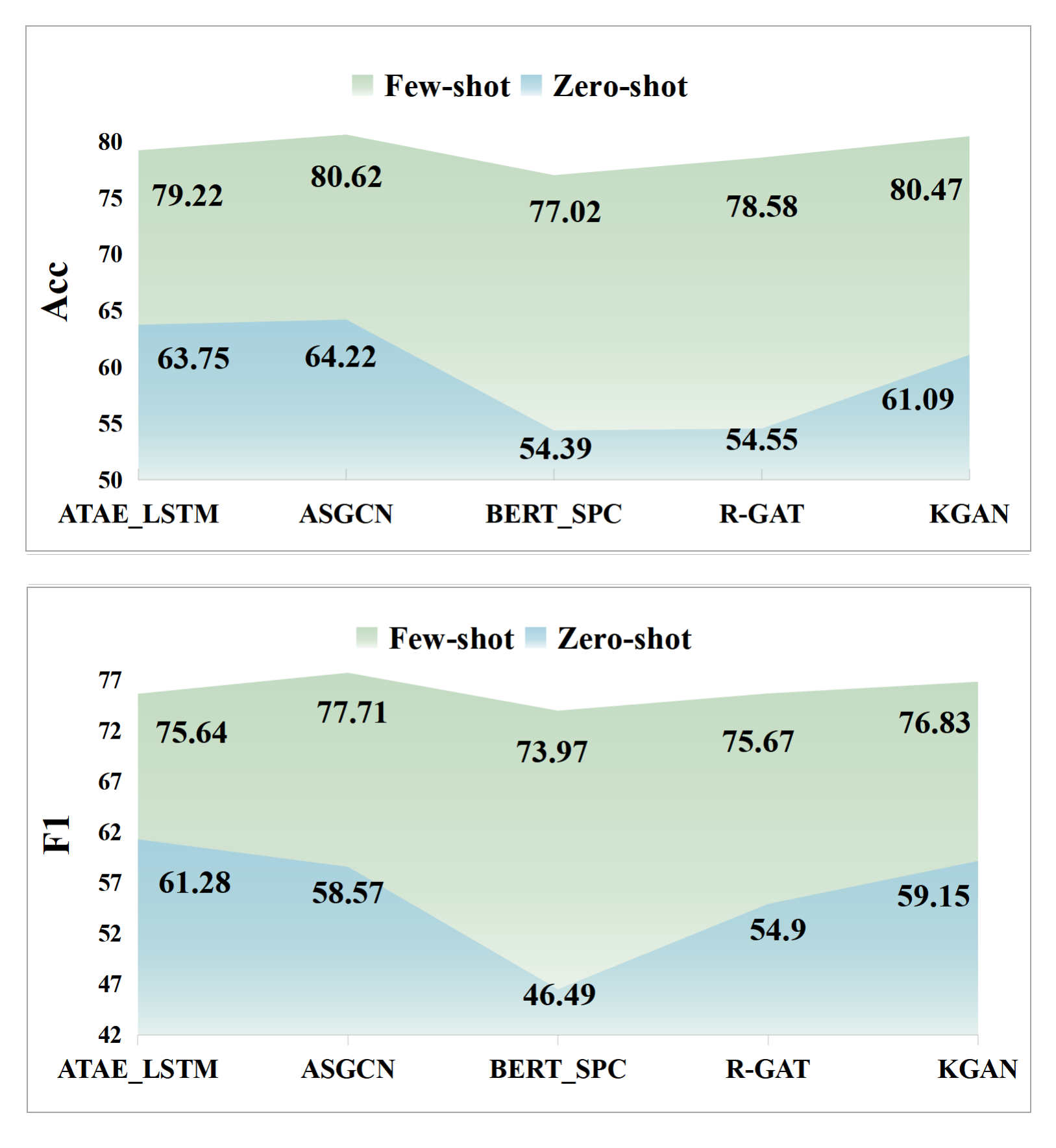}
    \caption{\textbf{Comparison of few-shot and zero-shot ITAT prompts in IDG.} We report the results on Laptop14 benchmark.
    }
    \label{fig:example_IN}
\end{figure}

\paragraph{\textbf{Effect of ITAT prompt}} 
In the iterative generation prompt (ITAT) of IDG, we use the high-quality synthesis data selected by the discriminator as demonstrations to guide the data generation of LLM. By doing so, IDG can make full use of the self-reflection abilities of LLM to boost the diverse and quality of synthesis data. Here, we compare the full IDG with a simple alternative,  \textit{i.e.}, removing the high-quality demonstrations in the prompt. For simplicity, we denote the full ITAT prompt as ``Few-shot'' and the simple alternative as ``Zero-shot''. The contrastive results on Laptop14 benchmark are shown in Fig.~\ref{fig:example_IN}. As seen, comparing to ``Zero-shot'', our IDG with the full ITAT prompt achieves better and more stable performance, indicating that \textit{adding some high-quality demonstrations in the ITAT prompt is beneficial to generate more high-quality data}. 

\begin{figure}[t]
    \centering   
    \includegraphics[width=0.47\textwidth]{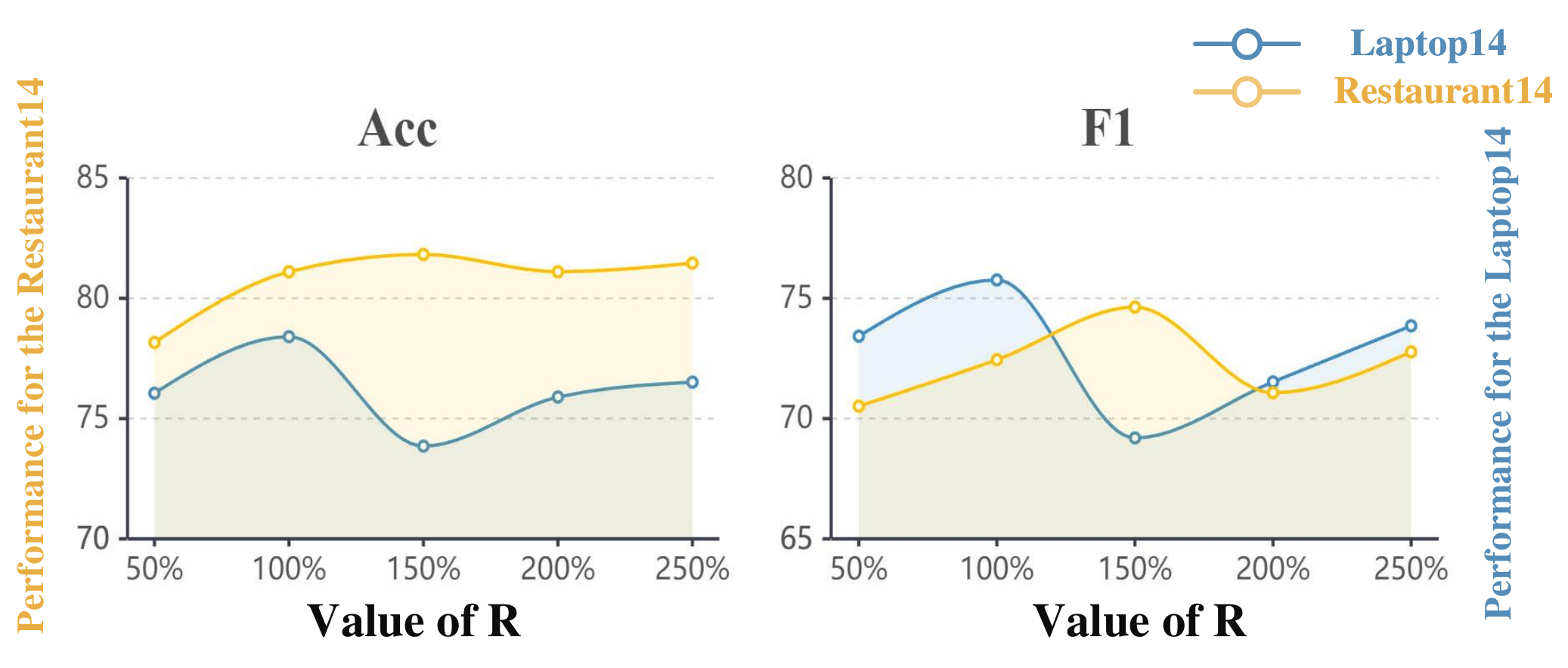}
    \caption{\textbf{Analysis on the number of generated data}. ``R'' denotes the ratio of the number of generated data relative to that of original training data. R-GAT is used as the baseline model in this experiment.
    }
    \label{fig:number}
\end{figure}

\paragraph{\textbf{Analysis of the number of generated data}} Here, we investigate the number of training data generated by IDG. Specifically, let $R$ be the number ratio of generated data relative to that of original training data, and we evaluate the performance of IDG with different $R$ ranging from 50\% to 250\%. Fig.~\ref{fig:number} illustrates the contrastive results of R-GAT on Laptop14 and Restaurant14 benchmarks. It can be found that the performance on both datasets shows a rising, falling, and then rising trend. With the increase in the amount of generated data, there will inevitably be more noisy samples in the generated data, which leads to performance degradation. However, with the generation of more reliable and stable quality samples, IDG brings performance improvements again. In general, these results show that \textit{more generated data does not always lead to better performance, \textit{i.e.}, data quality is more important than quantity.}

\section{Conclusion}
\label{sec:conclusion}
In this paper, we propose a systemic iterative data generation framework (IDG), which leverages the powerful abilities of LLMs to generate more high-quality labeled data. Starting from an unsupervised corpus, IDG first enforces the LLM to extract and expand the aspects and then designs an iterative LLM-based module to generate fluent and diverse labeled data. Lastly, IDG introduces a discriminator to filter the low-quality data. By doing so, IDG can effectively tackle the challenges of vanilla LLM-based data generation, \textit{i.e.}, LLMs are prone to hallucinations, leading to the unstable diversity and quality of synthesis data. Extensive experiments on four popular ABSA benchmarks upon fi baseline models show that the synthetic data generated by IDG can achieve comparable or even better performance against the original ground-truth data. Moreover, by combining the generated data and original data, IDG brings consistent and significant performance gains in all settings.

\section*{Acknowledgements}
This work was supported in part by the National Key Research and Development Program of China under Grant 2023YFC2705700, in part by the National Natural Science Foundation of China under Grants 623B2076, U23B2048, 62076186 and 62225113, and in part by the Innovative Research Group Project of Hubei Province under Grant 2024AFA017. The numerical calculations in this paper have been done on the supercomputing system in the Supercomputing Center of Wuhan University.

\ifCLASSOPTIONcaptionsoff
  \newpage
\fi

\bibliographystyle{IEEEtran}
\bibliography{TASLP.bib}

\end{document}